%% file: localprop_arxiv.tex
\documentclass{article} 
\usepackage{iclr2027_conference,times}

\input{math_commands.tex}

\usepackage{hyperref}
\usepackage{url}
\usepackage{graphicx} 
\usepackage{booktabs}
\usepackage{multirow}

\usepackage{pifont}

\title{LocalProp: Neuro-Localized\\ Memory-Efficient Backpropagation}

\iclrfinalcopy

\author{Diana-Nicoleta Grigore, Iuliana Georgescu, Radu Tudor Ionescu\thanks{Corresponding author: \texttt{raducu.ionescu@gmail.com}} \\
Department of Computer Science\\
University of Bucharest, Romania}

\usepackage{xspace}
\usepackage{tabularx}
\newcommand{\method}{LocalProp\xspace}
\newcommand{\xmark}{\ding{55}}

\begin{document}

\maketitle

\begin{abstract}
\vspace{-0.2cm}
The current deep learning training paradigm employs end-to-end backpropagation, regardless of the training stage, i.e.~pre-training or fine-tuning. However, backpropagating through the entire model is neither biologically plausible nor memory efficient, since learning inside the brain is highly localized. Therefore, we propose \method, a training procedure that locally updates the weights of a model. Our neuro-localized weight updates follow the ``pre-training then fine-tuning'' paradigm, where the pre-training is based on I-JEPA. After locally updating the  weights, a pruning operation is performed, followed by a short final fine-tuning phase. Pruning helps by sending the learning signal from higher blocks to lower blocks. We perform experiments on several datasets, including large-scale benchmarks such as ImageNet, and empirically show that \method reaches good performance at a fraction of GPU peak memory. By varying the number of jointly optimized blocks, we identify gradient-propagation span as a practical control over the accuracy--memory trade-off. 
\end{abstract}

\setlength{\abovedisplayskip}{2.4pt}
\setlength{\belowdisplayskip}{2.4pt}
\setlength{\abovedisplayshortskip}{2.4pt}
\setlength{\belowdisplayshortskip}{2.4pt}

\vspace{-0.1cm}
\section{Introduction}
\vspace{-0.1cm}

The development arc of the computer vision domain has been defined by the drive to learn increasingly robust and generalizable visual representations \citep{Dosovitskiy2021ViT, Liu_2021_ICCV}. 
To build and deploy state-of-the-art approaches for tasks such as detection \citep{carion2020end}, segmentation \citep{kirillov2023segment}, or 3D reconstruction \citep{mildenhall2020nerf}, the community has converged towards a multi-stage recipe: large-scale self-supervised pre-training \citep{caron2021dino, chen2020simple, grill2020byol, he2022masked, he2020moco}, followed by a task-specific supervised fine-tuning, and optionally concluding with structured pruning \citep{han2015learning,li2016pruning} or distillation~\citep{hinton2015distilling, romero2015fitnets, zhang2019byot, zhao2023cumulative} to meet inference constraints. Yet, across these stages, optimization relies on end-to-end backpropagation \citep{rumelhart1986learning}. Although undeniably successful, the global backward pass introduces obvious bottlenecks: it couples all layers under a single unified objective and typically requires storing or recomputing intermediate activations across the entire network depth \citep{chen2016training}.
\begin{figure}[tbp]
    \centering
    \begin{minipage}[t]{0.48\linewidth}
        \centering
        \includegraphics[
            width=\linewidth,
            keepaspectratio
        ]{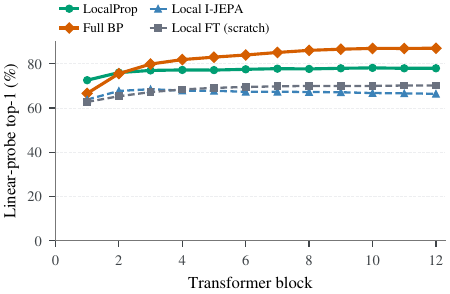}
        \vspace{-0.6cm}
        \caption{Representation development across ViT-S depth. Linear-probe top-1 accuracy at each transformer block for local I-JEPA pre-training, \method, full backpropagation fine-tuning, and local fine-tuning from scratch on CIFAR-10. \method rapidly improves feature separability in early blocks but then plateaus, while the full-BP upper bound continues to improve with depth.}
        \label{fig:credit}
    \end{minipage}\hfill
    \begin{minipage}[t]{0.48\linewidth}
        \centering
        \includegraphics[
            width=\linewidth,
            keepaspectratio
        ]{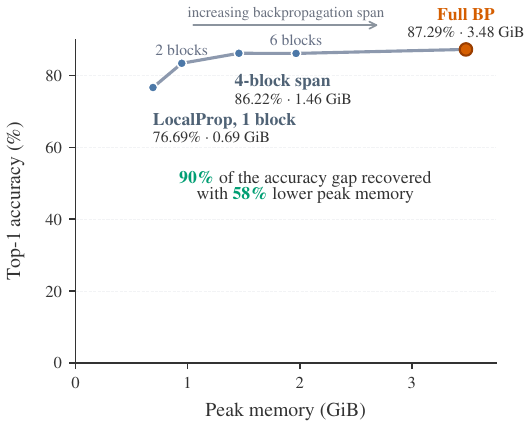}
        \vspace{-0.6cm}
        \caption{Accuracy-memory trade-off for CIFAR-10 fine-tuning as the local backpropagation window increases. Each point reports top-1 accuracy and peak allocated GPU memory when jointly optimizing 1, 2, 4, 6 blocks via \method vs.~12 blocks (full backpropagation). The four-block setting achieves 86.22\% accuracy at 1.46 GiB, using 58\% less memory than full backpropagation (87.29\% at 3.48 GiB) for a 1.07 percentage-point difference.}
        \label{fig:cifar10_backprop_budget}
    \end{minipage}
    \vspace{-0.4cm}
\end{figure}

\vspace{-0.1cm}
However, end-to-end backpropagation is not biologically plausible. Neural circuits in the brain do not freeze their forward processing to wait for explicit, global error signals over an entire architecture~\citep{lillicrap2020backprop}. Instead, learning and structural adaptation inside a living brain (\textit{in vivo}) are fundamentally decoupled and highly localized. Recent studies reveal that individual biological neurons simultaneously employ multiple distinct synaptic plasticity rules across different dendritic compartments, functioning entirely independently of a backward pass~\citep{wright2025distinct}. Furthermore, the biological counterpart of neural network pruning, experience-dependent synapse elimination in the adult brain, can be predicted from local synaptic co-activity and timing related to postsynaptic output ~\citep{hedrick2024local}. Inspired by these findings, we propose treating gradient propagation not as a strict requirement, but as a flexible design dimension.

\vspace{-0.1cm}
Several learning methods that do not use backpropagation have recently been proposed. The Forward-Forward algorithm \citep{hinton2022forward} uses two forward passes to update the weights of the model, while 
NoProp \citep{li2026noprop} trains the network by learning a local denoising process. However, previous works \citep{hinton2022forward, li2026noprop} tested their methods on small neural network architectures (e.g.~Forward-Forward \citep{hinton2022forward} uses three hidden layers) and small datasets (e.g.~Forward-Forward \citep{hinton2022forward} was tested on MNIST \citep{lecun2010mnist} and CIFAR-10 \citep{krizhevsky2009learning}, while NoProp \citep{li2026noprop} added CIFAR-100 \citep{krizhevsky2009learning}). Moreover, the aforementioned works are employed only during supervised training, but do not directly address the complete pipeline of self-supervised pre-training, supervised fine-tuning, and structured pruning considered here.



\vspace{-0.1cm}
We propose \method, a minimal-backpropagation framework for vision architectures that applies localized optimization during both pre-training and fine-tuning and extends the same gradient-locality constraint to pruning and recovery. Our approach unifies these stages under a constrained gradient budget. First, we decouple the network into isolated modules (transformer blocks or convolutional parameter groups) and optimize them using local error signals, detaching gradients at module boundaries. This allows us to circumvent the memory costs of standard full-depth backpropagation during large-scale self-supervised pre-training and task-specific fine-tuning. Second, rather than relying on a global loss controller for model compression, we extend this localized methodology to structural pruning. By ablating candidate components and evaluating their effect using a classifier, our method is motivated by the biological mechanism of experience-dependent synapse elimination. 
We emphasize that, while localized, block-wise learning has been studied previously, its viability across the full life-cycle of modern vision models, spanning pre-training, fine-tuning, and pruning has not been extensively explored.   

\vspace{-0.1cm}
We evaluate our \method framework across the entire model life-cycle on high-resolution benchmarks, such as ImageNet \citep{deng2009imagenet}, as well as CIFAR-10 and CIFAR-100 \citep{krizhevsky2009learning}. We employ two types of vision architectures: convolutional networks and transformers. We apply \method in the  pre-training stage, by modifying the I-JEPA \citep{assran2023self} algorithm to local updates. We compare the representations learned at sub-network level isolated by our local-gradient constraints against standard end-to-end approaches and same-size counterparts trained from random initialization, showing that our approach preserves critical structural priors and obtains highly competitive performance. 
Furthermore, we conduct a hardware efficiency analysis to demonstrate that confining optimization strictly to active blocks brings substantial benefits. As illustrated in Figure~\ref{fig:cifar10_backprop_budget}, a four-block \method reaches strong performance at a fraction (58\% less memory) of the peak GPU memory when compared to the full back-propagation algorithm.

\vspace{-0.1cm}
In summary, our study makes the following main contributions:
\begin{itemize}
    \item \vspace{-0.15cm} We show that self-supervised pre-training, supervised fine-tuning, and structured pruning can all be performed under the same locality constraint.
    \item \vspace{-0.08cm} We score attention heads or convolutional channels to obtain effective pruning paradigms. The obtained models outperform their same-size from-scratch counterparts, indicating that local pre-training and pruning preserve useful structure rather than merely selecting a smaller architecture.
    \item \vspace{-0.08cm} By restricting the backward graph, block-wise fine-tuning reduces peak memory per optimization step relative to full fine-tuning. These per-step savings are largest at ImageNet resolution, where avoiding full-depth activation storage substantially lowers memory usage.
\end{itemize}

\begin{figure}[t]
\centering
\includegraphics[width=\linewidth]{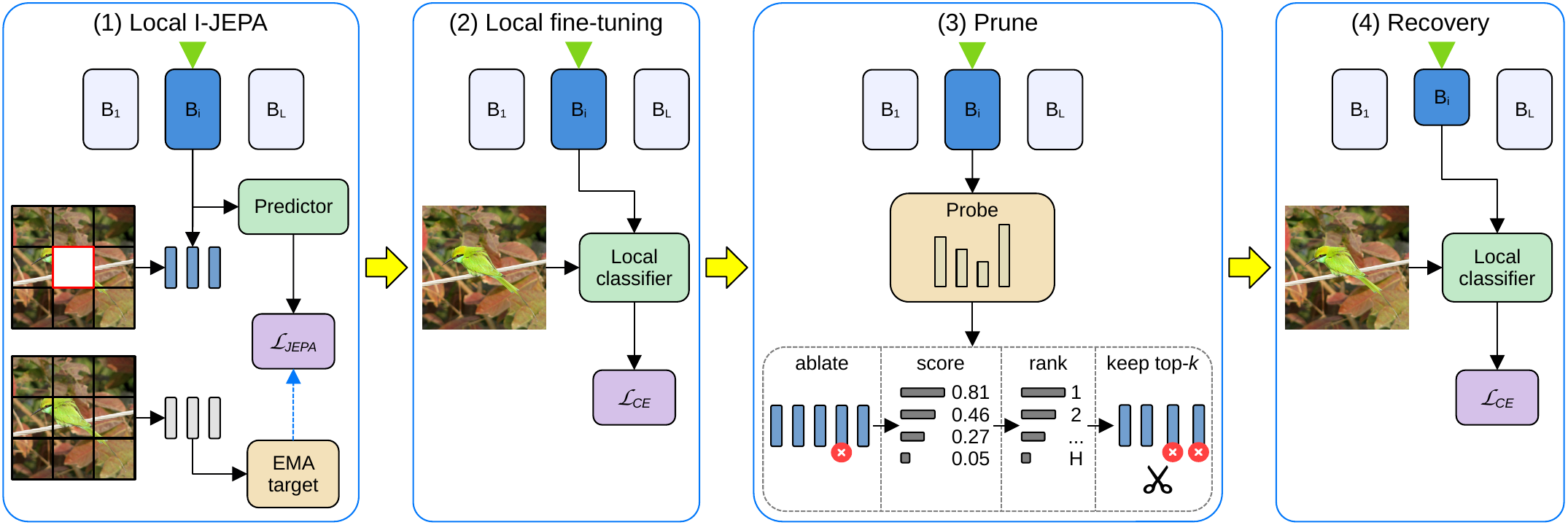}
\vspace{-0.6cm}
\caption{Overview of the \method training pipeline. Our method applies local optimization throughout I-JEPA pre-training, supervised fine-tuning, structured pruning, and local recovery. Only the selected block is optimized and scored at each step, avoiding end-to-end backpropagation. During pruning, downstream task performance is used to score lower-level components, allowing information from higher blocks to guide structural decisions without backpropagation. Best viewed in color.}
\label{fig:main_figure}
\vspace{-0.3cm}
\end{figure}

\vspace{-0.1cm}
\section{Related Work}
\vspace{-0.1cm}

We position \method with respect to three closely related directions: localized optimization, multi-stage vision training, and pruning strategies. While previous works have investigated these domains in isolation, \method demonstrates that a single, restricted-gradient budget can successfully unify them under one cohesive paradigm.

\vspace{-0.1cm}
\noindent \textbf{Localized optimization.}
Modern neural network training relies on backpropagation \citep{lecun2015deep, rumelhart1986learning, schmidhuber2015deep}. Despite its success, it brings substantial activation-memory costs and update locking, requiring error signals to propagate through the network in a way that does not closely reflect biological learning mechanisms \citep{lillicrap2016random, lillicrap2020backprop}. 
This tension has motivated learning rules that reduce, replace, or localize global backpropagation \citep{lee2015difference,lillicrap2016random, nokland2019training}. The \emph{Forward-Forward} algorithm \citep{hinton2022forward} replaces the backward pass with local goodness objectives. This paradigm has been further expanded to convolutional architectures via channel-wise competitive learning \citep{papac2024convolutional} and prototypical representations \citep{sarode2026hyperspherical}. Similarly, \emph{NoProp} \citep{li2026noprop} focuses on training networks by assigning local denoising objectives to individual blocks, while \emph{Greedy InfoMax} \citep{lowe2019end} has each layer independently learn to preserve the most salient structural information from the layer directly before it. These methods show that useful representations can emerge even when learning signals are not shared by the whole network. 

\vspace{-0.1cm}
Other approaches retain some gradient-based optimization, choosing to restrict the scope over which gradients are propagated. Local-error methods attach auxiliary objectives to group representations \citep{nokland2019training}, while \emph{Decoupled Greedy Learning} splits models into separately optimized modules to reduce update locking and enable more parallel training \citep{belilovsky2020decoupled}. \emph{SEDONA} \citep{pyeon2021sedona} further searches for effective network partitions to be locally trained, highlighting that the design of block-wise learning systems in itself is not trivial. \citet{siddiqui2024blockwise} train ResNet-50 blocks with local Barlow Twins objectives, while AugLocal \citep{ma2024scaling} uses auxiliary networks constructed from selected downstream layers to improve supervised local learning. These works complement DGL by demonstrating the scalability of local objectives in self-supervised and supervised settings. 

\vspace{-0.1cm}
\method follows this line of work, but takes a middle ground aimed at practical vision models. We do not eliminate backpropagation entirely and instead ask how far a vision model life-cycle can be carried out when backpropagation is confined to small local regions of the network. In our setting, gradients are restricted to the currently optimized block, while the surrounding computation is detached. This preserves the convenience of standard full-depth optimization, but avoids the memory and dependency cost.

\vspace{-0.1cm}
\noindent \textbf{Multi-stage vision pipelines.}
State-of-the-art vision models commonly rely on large self-supervised pre-training, followed by supervised adaptation. Contrastive methods \citep{chen2020simple, he2020moco} learn representations by comparing global network outputs across augmented views, while self-distillation approaches \citep{caron2021dino,grill2020byol} avoid explicit negative pairs. Masked-image methods instead predict missing content, as in Masked Autoencoders (MAE) \citep{he2022masked, hondru2025masked,madan2024cl}. Building upon this, Joint-Embedding Predictive Architectures (I-JEPA) \citep{assran2023self} avoid pixel-level reconstruction, training with the objective of predicting the latent representations of target patches from a visible context region. Once pre-training is complete, the foundational representations are specialized for downstream tasks through supervised fine-tuning \citep{yosinski2014transferable}. To meet the hardware constraints of deployment, this pipeline sometimes concludes with pruning \citep{han2015learning,li2016pruning} or distillation \citep{grigore2025weight} to eliminate excess architectural capacity. 

\vspace{-0.1cm}
These stages are commonly implemented using uninterrupted, end-to-end global backward passes for optimization. While prior localized learning methods attempt to remove this dependency, they are almost exclusively evaluated in isolation, addressing either self-supervised pre-training or adaptation, but rarely both. We instead consider the full training pipeline under a limited gradient budget, applying localized optimization from I-JEPA pre-training, through downstream fine-tuning, pruning, and recovery.

\vspace{-0.1cm}
\noindent \textbf{Pruning.}
Standard pruning methods remove complete channels, attention heads, or other components according to an importance criterion. Common approaches estimate this saliency using magnitude or loss-sensitivity \citep{li2016pruning, molchanov2019importance}, while dependency-based methods model couplings across layers \citep{fang2023depgraph}. In vision transformers, pruning is usually followed by end-to-end fine-tuning to recover performance \citep{yu2022unified}. Our framework uses gradient-free ablation scoring to assess structural importance through downstream classification performance. By ablating a channel and observing the drop in the downstream objective, we allow information from the network's final output to guide structural decisions. After pruning, recovery uses block-local supervised updates, while keeping the pruning masks enforced. This allows us to evaluate structured pruning and recovery under the same gradient-locality constraint used during pre-training and fine-tuning.

\vspace{-0.1cm}
\section{Method}
\label{sec:method}
\vspace{-0.1cm}   
            
\noindent \textbf{Overview.}
In Figure \ref{fig:main_figure}, we illustrate the entire pipeline of \method, composed of four stages: local I-JEPA pre-training, local supervised fine-tuning, pruning, and another short round of local fine-tuning to recover after pruning.

\vspace{-0.1cm}
\noindent \textbf{Local I-JEPA pre-training.}
We pre-train the neural model using a local variant of I-JEPA~\citep{assran2023self}. 
For each training image, we draw one augmented view and partition it into non-overlapping patches. For each image, we then sample one context mask and four target masks as sets of patch indices. The context mask determines which patch tokens remain visible to the online encoder, while the target masks specify the latent representations that must be predicted.

\vspace{-0.1cm}
At a selected neural block $i$, the online encoder processes the visible context only up to that depth. The target encoder processes the complete, unmasked token sequence through the same block and is evaluated without gradients. Its representations at the masked target positions provide the prediction targets. A block-specific predictor maps the context embeddings into a predictor space, appends learned masked tokens at the target positions, adds positional information, and predicts the target-encoder embeddings for those positions. For example, in vision transformers, 
each transformer block has its own predictor of exactly one transformer layer. 

\vspace{-0.1cm}
For any target token $t$, the local pre-training objective at depth $i$ is the Smooth $L_1$ distance between the predictor output $z_t^{(i)}$ and the detached target-encoder representation $h_t^{(i)}$:
            \begin{equation}
            \mathcal{L}^{(i)}_{\mathrm{JEPA}} =
            \frac{1}{|\mathcal{T}|}
            \sum_{t \in \mathcal{T}}
            \mathrm{SmoothL1}\!\left(z_t^{(i)}, \mathrm{sg}\!\left[h_t^{(i)}\right]\right),
            \end{equation}
where $\mathcal{T}$ denotes the sampled target patch positions and $\mathrm{sg}$ denotes the stop-gradient operation. Image labels are not used during pre-training.

\vspace{-0.1cm}
The defining constraint enabling \textit{local} backpropagation training is that only one transformer block is optimized at a time. At each update, the training loop chooses an active block using a round-robin schedule and computes the I-JEPA objective at that depth. Earlier blocks are evaluated without gradients, and blocks following it are not evaluated. Consequently, gradients are restricted to the active block and its corresponding predictor. The shared patch embedding is updated during the processing of the first transformer block. Similarly to I-JEPA, the target block and target patch embedding are updated by exponential moving average (EMA) from their online counterparts rather than by gradient descent.

\vspace{-0.1cm}
\noindent \textbf{Local supervised fine-tuning.}
After self-supervised pre-training, we adapt the encoder to a labeled downstream dataset using block-local supervised losses. At each optimizer step, one single neural block is selected and optimized for a few updates, after which the schedule advances to the next block in round-robin order. The output patch tokens are average-pooled, and a block-specific linear classifier predicts the target. For a selected block $i$, the base fine-tuning objective is the cross-entropy with label smoothing:
            \begin{equation}
            \mathcal{L}^{(i)}_{\mathrm{sup}} =
            \mathrm{CE} \left(
             c_i\left(\mathrm{pool}\left(h^{(i)}\right)\right), y
            \right),
            \end{equation}
where $h^{(i)}$ is the token sequence emitted by block $i$, $c_i$ is its local classifier, and $y$ is the ground-truth class.

\vspace{-0.1cm}
Only the active block and its corresponding classifier are optimized during an update. The shared patch embedding is additionally optimized during the update burst assigned to the first transformer block. Applying the same training procedure to a randomly initialized encoder gives the local supervised-from-scratch baseline. Final predictions use the classifier attached to the last transformer block.
            

\vspace{-0.1cm}
\noindent \textbf{Local structured pruning.}
Pruning begins from a local-wise fine-tuned neural model and processes candidate blocks from deepest to shallowest. Before pruning, we load or fit a lightweight classifier on the pooled output tokens of the final block, while keeping the backbone frozen. The classifier provides a common downstream objective for scoring components throughout the network. For a candidate component $u$ in block $i$, we temporarily suppress its output and evaluate the resulting final classification loss. Let $\mathcal{L}_{\mathrm{final}}$ denote the baseline loss and let $\mathcal{L}^{(i)}_{\mathrm{final},-u}$ denote the loss after suppressing component $u$ in block $i$. We define its non-negative importance score as: 
            \begin{equation}
            s_u^{(i)}
            =
            \max\left(
            0,
            \mathcal{L}^{(i)}_{\mathrm{final},-u}
            -
            \mathcal{L}_{\mathrm{final}}
            \right).
            \end{equation}
A large score indicates that removing the component degrades the final prediction. Scoring requires a forward pass through the complete encoder, but does not require gradient computation or backpropagation. For a partial-retention budget, we keep the highest-scoring components. When the prescribed budget removes an entire branch, all its components are removed and no ranking is required. 

\vspace{-0.1cm}
For attention-head pruning, suppressing a head removes both sides of its computation. We zero the corresponding rows of the combined query, key, and value projection and the matching input columns from the attention output projection. When the desired number of heads $H_{\mathrm{keep}}>0$, the surviving output-projection columns are rescaled by $\sqrt{H/H_{\mathrm{keep}}}$, where $H$ is the
original number of heads. This approximately preserves the attention-output scale after pruning and limits disruption to the residual stream.

\vspace{-0.1cm}
After pruning a block, we fit a classifier to its local representation and optimize only that block against the corresponding local supervised objective. The accumulated structured mask remains enforced in recovery mode. After processing all candidates, we perform a final local cleanup stage over the complete encoder under the final mask, using the same schedule described in the fine-tuning stage.

            \begin{table}[t]
                        \caption{Effect of \method training phases across architectures and datasets. Top-1 accuracy on CIFAR-10, CIFAR-100, and ImageNet-1K for ViT-S and ResNet-50 under multiple configurations of local pre-training, fine-tuning, pruning, and recovery. Reduced-size variants use the same parameter budget as the pruned models. Local pre-training improves subsequent local fine-tuning, while recovery restores much of the performance lost during pruning. 
            }
            \vspace{0.1cm}
            \label{tab:full_pipeline}
            \centering
            \scriptsize
            \setlength\tabcolsep{1.8pt}
            {
            \begin{tabular}{@{}lcccccccccc@{}}
            \toprule
            \multirow{2.5}{*}{Setting}
            & \multirow{2.5}{*}{\shortstack{Local\\I-JEPA}}
            & \multirow{2.5}{*}{Pruning}
            & \multirow{2.5}{*}{Recovery}
            & \multirow{2.5}{*}{\shortstack{Param.\\factor}}
            & \multicolumn{3}{c}{ViT-S}
            & \multicolumn{3}{c}{ResNet-50} \\
            \cmidrule(lr){6-8}
            \cmidrule(lr){9-11}
            &
            &
            &
            & & CIFAR-10 & CIFAR-100 & ImageNet-1K & CIFAR-10 & CIFAR-100 & ImageNet-1K \\
            \midrule
            
            Local FT (scratch) &
             \xmark & \xmark &  \xmark  &  $1.2\times$ & 68.63 & 42.53 & 33.29 & 87.75 & 65.31 & 47.36 \\
            \method (no pruning) &
            \checkmark & \xmark & \xmark & $1.2\times$  &76.69 & 50.53 & 37.12 & 89.79 & 67.07 & 55.75 \\
            \midrule
            \method (no recovery) &
            \checkmark & \checkmark  & \xmark & $1.0\times$  &52.21 & 27.62 & 24.96 & 88.23 & 52.94 & 37.83 \\
            \method &
            \checkmark & \checkmark  & \checkmark & $1.0\times$  &75.40 & 47.48 & 34.81 & 88.84 & 67.26 & 52.19 \\
            \midrule
            Local I-JEPA probe & \checkmark & \xmark & \xmark & $1.2\times$ & 49.46 & 19.16 & 8.27 & 32.80 & 11.09 & 11.89 \\
            Scratch (reduced) &
            \xmark & \xmark & \xmark & $1.0\times$ & 66.85 & 39.17 & 21.48 & 85.87 & 62.71 & 47.68\\
            \method &
            \checkmark & \xmark & \xmark  & $1.0\times$ & 70.71 & 44.00 & 6.09 & 87.59 & 65.60  & 1.49 \\
            \midrule
            Full BP (reference) &
            -- & -- & -- & $1.0\times$ &
            87.29 & 85.70 & 76.87 & 97.40 & 85.20 & 76.13 \\
            \bottomrule
            \end{tabular}
            }
            \end{table}

\vspace{-0.1cm}
\section{Experiments}
\label{sec:experiments}
\vspace{-0.1cm}
\subsection{Datasets}
\vspace{-0.1cm}
We evaluate \method on CIFAR-10, CIFAR-100 \citep{krizhevsky2009learning}, and ImageNet-1K \citep{deng2009imagenet}, three general-purpose benchmarks covering increasingly challenging settings, from a low-resolution of $32 \times 32$ pixels to natural images of $224\times224$ pixels. CIFAR-10 and CIFAR-100
each contain $50,000$ training and $10,000$ test images, while ImageNet-1K contains $1,281,167$ training and $50,000$ validation images across $1,000$ classes. We include ImageNet-1K as an explicit scalability test, since several recent approaches that avoid or localize backpropagation are evaluated primarily on small, CIFAR-scale benchmarks. This allows us to study whether local optimization remains effective when both dataset scale and input resolution increase substantially.

\vspace{-0.1cm}
\subsection{Implementation Details}
\vspace{-0.1cm}

As underlying models, we consider ViT-S/16, containing 12 blocks, 384-dimensional embeddings, and 6 attention heads, as well as ResNet-50, which we divide into 14 local groups following its bottleneck structure. For the convolutional network, we increase the capacity of the later stages using bottleneck width multipliers $[1.0, 1.0, 1.5, 1.5]$, corresponding to internal widths $[64,128,384,768]$. ViT-S operates on $224\times224$ inputs and is locally pre-trained on ImageNet-1K for $600$ epochs using AdamW, an effective batch size of $4096$, and a cosine learning rate schedule starting at $10^{-4}$. We use one context and four target masks, and update one transformer block per step in round-robin order. For the ResNet experiments on CIFAR-10/100, local I-JEPA pre-training lasts $200$ epochs with batch size $128$, groups are selected using round-robin, and optimized using detached inter-group activations. All pre-training experiments use BF16 and EMA target encoders.

\vspace{-0.1cm}
Local supervised fine-tuning uses independent classifiers and losses for each block or group. The transformer is fine-tuned for 120 epochs with AdamW and a learning rate of $5\times10^{-5}$ for a batch size of $64$, updating one block at a time, whereas the convolutional network is trained for 500 epochs with detached group boundaries. An epoch denotes one pass through the training dataset overall, rather than a separate pass per block. In pruning, we use late-block attention-head pruning and channel pruning. Component importance is scored using only the training split of the respective dataset. Recovery employs the same local supervised objectives, while enforcing structural masks throughout optimization. Reported active parameter counts exclude masked weights. Tensors are not physically compacted in these experiments.

\begin{table}[t]
\centering
\scriptsize
\caption{Peak allocated memory for ViT-S is measured on an RTX 3090. Full updates backpropagate
through the complete encoder, whereas local updates optimize one
transformer block per step.
}
\label{tab:memory_runtime_local_training}
\vspace{0.1cm}
\begin{tabular}{llccc}
\toprule
Input & Setting & Peak memory & Memory reduction & Top-1 accuracy \\
\midrule
\multirow{2}{*}{CIFAR-10}
& Full            & 3.48 GiB & 0.0\%  & 87.29\% \\
& Local (1 block) & 0.69 GiB & 80.2\% & 76.69\% \\
\midrule
\multirow{2}{*}{ImageNet-1K}
& Full            & 3.40 GiB & 0.0\%  & 76.87\% \\
& Local (1 block) & 0.74 GiB & 78.2\% & 37.12\% \\
\bottomrule
\end{tabular}
\end{table}

\begin{table}[t]
\caption{Comparison with alternatives to full backpropagation. Top-1 accuracy of \method, NoProp, Greedy InfoMax, Forward-Forward, and DGL on CIFAR-10, CIFAR-100, and ImageNet-1K using ResNet-50 and ViT-S backbones. Results unavailable in the cited works are obtained from our reproductions. NV (no variant) denotes a backbone variant unavailable in the evaluated implementation. EB (exceeding budget) denotes a configuration whose full-run training time exceeded our per-run budget of 72 GPU-hours (3 days) on an NVIDIA RTX 3090.}
\label{tab:local_baselines}
\vspace{0.1cm}
\centering
\scriptsize
\setlength{\tabcolsep}{1.9pt}
\begin{tabular}{lccccccc}
\toprule
\multirow{2.5}{*}{Method}
& \multirow{2.5}{*}{\shortstack{Parameter\\factor}}
& \multicolumn{3}{c}{ResNet-50}
& \multicolumn{3}{c}{ViT-S} \\
\cmidrule(lr){3-5}
\cmidrule(lr){6-8}
&
& CIFAR-10 & CIFAR-100 & ImageNet
& CIFAR-10 & CIFAR-100 & ImageNet \\
\midrule
\method (no pruning)
& $1.2\times$
& 89.79 & 67.07 & 55.75
& 76.69 & 50.53 & 37.12 \\

\method
& $1.0\times$
& 88.84 & 67.26 & 52.19
& 75.40 & 47.48 & 34.81 \\
\midrule
NoProp \citep{li2026noprop}
& $1.2\times$
& 83.91 & 52.02 & EB
& NV & NV & NV \\

Greedy InfoMax \citep{lowe2019end}
& $1.2\times$
& 72.24 & 41.02 & 3.11
& 42.58 & 15.55 & 18.81 \\

FF \citep{papac2024convolutional, sarode2026hyperspherical}
& $1.2\times$
& 78.11 & 51.23 & 25.00
& NV & NV & NV \\

DGL \citep{belilovsky2020decoupled}
& $1.2\times$
& 83.40 & 57.60 & 57.14
& 78.36 & 47.02 & 36.88 \\
\bottomrule
\end{tabular}
\end{table}

\begin{table}[t]
\centering
\begin{minipage}[t]{0.48\linewidth}
    \vspace{0pt}
    \caption{Ablation of interface objective for ViT-S on CIFAR-10. Top-1 accuracy obtained with different detached interface targets. Representation-level cosine matching performs best, with little difference between neighboring and final block targets.}
    \label{tab:interface-objective-detached-target}
    \vspace{0.1cm}
    \centering
    \small
    \setlength{\tabcolsep}{4pt}
    \renewcommand{\arraystretch}{1.1}
    \fontsize{8.5}{9}\selectfont{
    \begin{tabularx}{\linewidth}{
        @{}>{\raggedright\arraybackslash}Xcc@{}
    }
        \toprule
        Interface objective & Target & C10 \\
        \midrule
        Pooled cosine
        & \shortstack{Neighbor block} & 77.20 \\
        Pooled cosine
        & Final block & \textbf{77.25} \\
        Token cosine with delay/ramp
        & Final block & 77.08 \\
        Logit KL with delay/ramp
        & Final block & 76.23 \\
        \bottomrule
    \end{tabularx}
    }
\end{minipage}\hfill
\begin{minipage}[t]{0.48\linewidth}
    
    \caption{Ablation of pruning strategies for ViT-S. Comparison of uniform pruning, global head pruning, block dropping and late-block head pruning on CIFAR-10 and CIFAR-100.}
    \label{tab:pruning-placement-summary}
    \vspace{7pt}
    \vspace{0.1cm}
    \centering
    \small
    \setlength{\tabcolsep}{2pt}
    \renewcommand{\arraystretch}{1.1}
    \fontsize{8.5}{9}\selectfont{
    \begin{tabularx}{\linewidth}{
        @{}>{\raggedright\arraybackslash}Xccc@{}
    }
        \toprule
        Setting & \#Params
        & \shortstack{C10}
        & \shortstack{C100} \\
        \midrule
        \method (no pruning)
        & 21.66M & 76.69 & 50.53 \\
        \midrule
        Global head pruning
        & 16.94M & 69.02 & 41.49 \\
        Late-block head pruning
        & 16.93M & \textbf{75.40} & \textbf{47.48} \\
        Every third block removed
        & 14.57M & 73.55 & 43.88 \\
        Uniform 50\% retention
        & 11.03M & 66.55 & 37.35 \\
        \bottomrule
    \end{tabularx}
    }
\end{minipage}
\end{table}

\vspace{-0.1cm}
\subsection{Results}
\vspace{-0.1cm}

\noindent \textbf{\method performance.} In Table \ref{tab:full_pipeline}, we summarize the performance of \method training phases: local pre-training, local adaptation, and pruning. The benefit of local self-supervised pre-training strongly depends on the architecture. For ViT-S, local I-JEPA pre-training provides a consistently better initialization for local fine-tuning, improving over local training from scratch on CIFAR-10, CIFAR-100, and ImageNet-1K. In contrast, ResNet-50 already performs strongly under purely local supervised training, and the contribution of pre-training is most evident on ImageNet-1K, suggesting that convolutional inductive biases make the architecture less dependent on the self-supervised initialization on smaller datasets. We offer a full-backpropagation comparison from existing published results \citep{chen2022when, mo2024connecting, torchvision_resnet50}.

\vspace{-0.1cm}
Our pruning strategy has slightly different effects on the two architectures. For ViT-S, our main late-block pruning configuration reduces the active parameter count by 21.8\%. For ResNet-50, channel pruning removes 24.3\%. Despite similar parameter-reduction ratios, the architectures respond differently to pruning. ViT-S suffers a substantial immediate drop after pruning ($37.12\%$ to $24.96\%$ on ImageNet-1K), but the second supervised fine-tuning restores the accuracy toward that of the full-size model (from $24.96\%$ to $34.81\%$ on ImageNet-1K), indicating that the remaining blocks can locally reorganize after components are removed. ResNet-50 is robust immediately after pruning (drop from $89.79\%$ to $88.23\%$ on CIFAR-10), while the benefit of recovery is less uniform across datasets. This suggests that the usefulness of local recovery depends not only on the amount of pruning, but also on how redundancy is distributed across architectures and datasets. 

\vspace{-0.1cm}
Importantly, the recovered pruned models outperform their same-size counterparts (parameter factor $1.0\times$) initialized from scratch in every comparison, e.g.~$75.40\%$ vs.~$66.85\%$ for ViT-S on CIFAR-10, and $34.81\%$ vs.~$21.48\%$ for ViT-S on ImageNet-1K. Thus, their performance cannot be explained solely by the capacity of the resulting smaller architecture: useful information learned before pruning survives the structural modification and can be recovered using local updates. The advantage diminishes, but remains substantial ($75.40\%$ vs.~$70.71\%$ for ViT-S on CIFAR-10), when the same-size architecture (parameter factor $1.0\times$) is itself locally pre-trained with I-JEPA.


\vspace{-0.1cm}
\noindent \textbf{Backpropagation budget.} In Figure~\ref{fig:cifar10_backprop_budget}, we vary the number of jointly optimized ViT blocks from $1$ to $12$, treating gradient propagation as a tunable coordination budget. Larger windows improve accuracy, but require retaining more activations for backpropagation. The graph in Figure~\ref{fig:cifar10_backprop_budget} shows that most of the accuracy benefit of full backpropagation is recovered well before reaching the full model depth (12 blocks). A four-block window recovers $90\%$ of the accuracy gap between one-block training and full fine-tuning ($86.22\%$ vs.~$87.29\%$), while using $58 \%$ less peak memory. These results suggest that gradient propagation is not always a binary choice between local and global learning. Instead, the propagation span acts as a tunable coordination budget, with larger windows allowing optimization to coordinate across more blocks, while shorter windows reduce the cost of training. This could also be employed to train very deep transformers without splitting the model across GPUs, therefore decreasing the communication time while retaining the performance.

\begin{figure}[!t]
\centering
\includegraphics[width=0.72\linewidth]{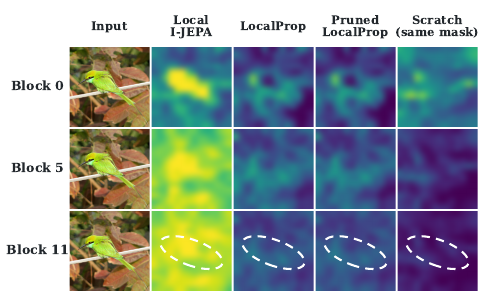}
\vspace{-0.2cm}
\caption{Feature visualizations on an ImageNet validation image. Feature magnitude is shown at transformer blocks $0$, $5$, and $11$ for local I-JEPA, \method, pruned \method, and the same-mask architecture trained from scratch. Colors are jointly normalized across methods within each block row, and the dashed contour marks the same object region in all final maps. \method develops a clearer object-aligned late-layer response than the scratch model, which is largely preserved after pruning.}
\label{fig:feature-visualization}
\end{figure}

\vspace{-0.1cm}
\noindent \textbf{Understanding local backpropagation.} The results shown in Figure~\ref{fig:credit} indicate that \method achieves strong linear separability in early blocks but saturates later, while full backpropagation continues to improve. The supplementary analysis shows positively aligned local and full-backpropagation gradients with similar magnitudes. These observations motivate our extended gradient-span experiments: although strict local training learns useful representations, independently optimizing each block limits how the model benefits from increasing depth. The qualitative feature maps in Figure \ref{fig:feature-visualization} show more spatially concentrated activations for \method than for local training from scratch, particularly in later blocks, while the pretrained variants retain clearer spatial structure.

\vspace{-0.1cm}
\noindent \textbf{Structured pruning and recovery.} The results in Table \ref{tab:pruning-placement-summary} show that pruning performance depends on where capacity is removed. At nearly the same active-parameter budget, concentrating head pruning in later ViT blocks substantially outperforms global head pruning, indicating that attention capacity in the earlier parts of the model is more sensitive to removal under local training. Dropping every third transformer block performs better than global head pruning, but still remains below late-block head pruning, suggesting that preserving network depth is preferable to eliminating entire blocks. Finally, aggressive uniform pruning produces the largest degradation, showing that compression quality depends on both the amount and structural location of removed capacity.

\vspace{-0.1cm}
\noindent \textbf{Comparison with backpropagation alternatives.}
Table \ref{tab:local_baselines} compares \method with approaches that restrict or avoid global backpropagation. The clearest trend is that the gap between \method and competing methods grows as the problem becomes more challenging. On CIFAR, several approaches retain reasonable performance, whereas the gaps over Greedy InfoMax and Forward-Forward become substantially larger on ImageNet-1K  suggesting that maintaining useful signals becomes more difficult as dataset scale and class diversity increase. DGL, which uses auxiliary prediction networks during training \citep{belilovsky2020decoupled}, remains competitive, with mixed results across settings.

\vspace{-0.1cm}
Across architectures, Greedy InfoMax degrades sharply when transferred from ResNet-50 to ViT-S, while \method remains effective for both convolutional and transformer backbones. The pruned \method models also remain competitive, while operating with roughly one quarter fewer active parameters. Overall, these results indicate that the advantage of our method is most pronounced in settings where local optimization is hardest.

\vspace{-0.1cm}
\noindent \textbf{Efficiency.}
The measurements reported in Tables \ref{tab:memory_runtime_local_training} and \ref{tab:backprop_cost_summary} show that the computational advantage of local optimization becomes more pronounced as input resolution increases. While localized updates already reduce memory at CIFAR scale, the gap becomes substantially larger on ImageNet, where full backpropagation must retain high-resolution activations throughout the entire network. This indicates that the main savings come from shortening the backward graph, making \method increasingly attractive as training cost scales with model and input size. The per-step savings quantify the benefit of optimizing one block rather than all of them. Total training time additionally depends on the update schedule and the accuracy target.

\vspace{-0.1cm}
\noindent \textbf{Interface objective.} We further explore whether locally optimized layers can communicate through a detached representation-matching signal, without restoring end-to-end backpropagation. Table \ref{tab:interface-objective-detached-target} shows that representation-level cosine matching performs best, with little difference between neighboring-block and final-block targets, while logit-level KL is less effective. These results suggest that additional information from cross-block supervision can help, but is not a major source of performance gain.



\begin{table}[t]
\caption{Training cost of block-wise local fine-tuning vs.~end-to-end fine-tuning for ViT-S/16. Backward operations are reported per optimization step, and VRAM is the peak allocated memory on an RTX 3090. GPU-hours are measured for block-wise ImageNet fine-tuning and full fine-tuning over 120 epochs.}
\label{tab:backprop_cost_summary}
\centering
\small
\setlength{\tabcolsep}{8pt}
\renewcommand{\arraystretch}{1.12}
\begin{tabular}{lcccc}
\toprule
{Method} &
{\#blocks} &
{TFLOPs} &
{GiB} &
{GPU-h} \\
\midrule
Local FT
  & $1/12$ 
  & $0.195$ 
  & $0.74$ 
  & $27.8$  \\
Full FT
  & $12/12$ 
  & $2.340$ 
  & $3.40$  
  & $75.1$  \\
\midrule
\textbf{Reduction}
  & \textbf{$12.0{\times}$}
  & \textbf{$12.0{\times}$}
  & \textbf{$4.6{\times}$}
  & \textbf{$47.3$} saved \\
\bottomrule
\end{tabular}
\end{table}

\vspace{-0.1cm}
\section{Conclusion}
\vspace{-0.1cm}

In this paper, we introduced \method, a novel minimal-backpropagation framework based on localized gradient updates. Our approach applies the same locality principle throughout self-supervised pre-training, supervised fine-tuning and structured pruning with recovery. We conducted experiments on CIFAR-10, CIFAR-100, and ImageNet-1K using transformers and convolutional networks, and considered different gradient-propagation spans and pruning settings. The results show that \method compares favorably with existing local optimization approaches, while substantially reducing peak training memory. Our analyses show that local objectives provide useful and comparably scaled learning signals, but miss out on the coordination across depth. These findings support viewing gradient propagation as a tunable coordination budget and are consistent with the biological motivation that effective plasticity can arise from locally available signals without network-wide information transport.

\subsection*{Funding}
This research is supported by the project ``Romanian Hub for Artificial Intelligence - HRIA'', Smart Growth, Digitization and Financial Instruments Program, 2021-2027, MySMIS no.~351416.

\subsection*{AI use statement}

We used generative AI tools for coding assistance and LaTeX formatting of the tables and figures. The authors reviewed all AI-assisted contributions and take responsibility for the final content of this work, including its code, text, claims, and reported results.

We did not use generative AI tools to generate synthetic data sets, help develop theoretical models or conceptual frameworks, formulate mathematical claims, provide critical ingredients for proving mathematical claims, assist in the writing of proofs, propose or refine hypotheses, design or provide feedback on research methodology or experiments, assist with translation, clean and reformat dataset, support qualitative and thematic data analysis, and interpret results. 

\subsection*{Ethics statement}

This work studies memory-efficient training of vision models using established benchmarks: CIFAR-10, CIFAR-100, and ImageNet-1K. Reducing memory requirements may improve access to model training on limited hardware. The resulting models may nevertheless inherit biases from their training data, and deployment in sensitive applications requires
additional assessment of fairness, privacy, and potential misuse.

\subsection*{Reproducibility statement}

Section~\ref{sec:method} describes the local learning objectives, block-selection schedule, and pruning and recovery procedures. Section~\ref{sec:experiments} provides the datasets, model configurations, and main training hyperparameters. Additional implementation details, experimental settings, and analyses are provided in the appendix. 

\bibliography{iclr2027_conference}
\bibliographystyle{iclr2027_conference}

\appendix
\section{Appendix}

\subsection{Additional Method Details}

\noindent \textbf{Local I-JEPA for ResNet-50.} 
We provide the architectural specifics of the ResNet-50 instantiation of local I-JEPA \citep{assran2023self}. Adapting the objective from vision transformers is not straightforward because, unlike the fixed token grid used by ViT, ResNet-50 repeatedly downsamples its feature maps and increases their channel dimensionality across stages. As a result, context and target masks defined in the input space must be aligned with a different spatial resolution at each local group, while predicting the full  target feature map would become increasingly expensive at deeper layers. CNN-JEPA \citep{kalapos2024cnnjepa} addresses these differences by aligning masks with  the encoder's spatial downsampling and using a lightweight predictor built from depthwise-separable convolutions. Inspired by these architectural choices, we resize the input-space context and target masks to the resolution of each local group and use a similar predictor. Our predictive objective is instantiated independently at multiple local groups under the \method gradient constraint.

We partition ResNet-50 into $14$ local groups. The first contains the convolutional stem and the first bottleneck. Each remaining bottleneck through Stage~3 forms an individual group, while the final group contains the complete Stage~4. We chose this split to avoid introducing increasingly small local modules. Each local group is followed by a $1{\times}1$ projection and a predictor composed of three residual depthwise-separable convolutional blocks.

 We use a resolution-agnostic group-level objective for the convolutional variant. Averaging over the target-mask region avoids requiring exact element-wise correspondence after repeated spatial downsampling, while $\ell_2$ normalization and cosine distance reduce sensitivity to stage-dependent feature magnitudes. Consequently, the objective matches the semantic content of the masked region, whereas the ViT objective retains the finer token-wise Smooth L1 supervision:
\begin{equation}
\mathcal{L}^{(i)}_{\mathrm{JEPA}} =
1 -
\left\langle
\widehat{p}^{(i)},
\mathrm{sg}\!\left[\widehat{h}^{(i)}\right]
\right\rangle ,
\end{equation}
where $\widehat{p}^{(i)}$ and $\widehat{h}^{(i)}$ denote the normalized
predicted and target representations for group $i$, respectively, and
$\mathrm{sg}$ denotes stop-gradient.

\begin{table}[!t]
\caption{Key hyperparameters for Local I-JEPA pre-training.
The active transformer block or ResNet group is selected in
round-robin order. For ViT-S, the learning-rate entry denotes the initial learning rate before warmup and cosine decay. The three ResNet-50 rates
correspond to group $0$, groups $1$--$12$, and group $13$, respectively.}
\label{tab:ijepa_hparams}
\vspace{0.1cm}
\centering
\small
\setlength{\tabcolsep}{3pt}
\renewcommand{\arraystretch}{0.98}
\begin{tabular}{@{}lcc@{}}
\toprule
Hyperparameter & ViT-S/16 & ResNet-50 \\
\midrule
Pre-training epochs
& 600
& 200 \\

Optimizer
& AdamW
& AdamW \\

Learning rate
& $10^{-4}$
& $[1,3,.75]\!\times\!10^{-4}$ \\

Warmup
& 40 ep.
& 5 ep. \\

Weight decay
& $0.04$
& $10^{-4}$ \\

Batch size
& 4096 (effective)
& 128 \\

Precision
& BF16
& BF16 \\

EMA momentum
& $.996\!\rightarrow\!1$
& $.999\!\rightarrow\!1$ \\

Update schedule
& 1 block/step
& 1 group/step \\

\bottomrule
\end{tabular}
\end{table}

\noindent \textbf{Local supervised fine-tuning for ResNet-50.}
Unlike the round-robin ViT procedure, all $14$ ResNet groups are evaluated and updated on every mini-batch. The input to each group is detached from the preceding group, and its output feature map is globally average-pooled and passed to a local linear classifier over the original dataset classes.

The $14$ local losses are summed before backpropagation. Because activations are detached at every group boundary, each loss updates only its corresponding group and classifier. After adaptation, the local classifiers are discarded, and performance is evaluated by training a linear probe on the frozen final-group representations.

\noindent \textbf{Channel pruning and local recovery for ResNet-50.} For ResNet-50, channel importance is evaluated using a fixed linear classifier attached to the final network representation. Each candidate is temporarily ablated, and its importance is defined as the resulting increase in final classification loss. Channels with the highest scores are retained until the target width is reached. The resulting structured mask sets the removed convolutional filters and their associated batch-normalization parameters to zero. Recovery uses the same group-local supervised objective. The structured masks are applied to both parameters and gradients throughout optimization, ensuring that removed channels remain inactive.

\begin{table}[!t]
\caption{Effect of pruning order on ResNet-50.
CIFAR-10 top-1 accuracy after channel pruning and recovery. Both traversal orders use the same starting checkpoint, pruning budget, scoring procedure, and structured masked constraints, removing $24.29\%$ of the parameters. The main configuration processes layers from deepest to shallowest. 
}
\label{tab:pruning_order}
\vspace{0.1cm}
\centering
\small
\setlength{\tabcolsep}{4pt}
\renewcommand{\arraystretch}{1.1}
\begin{tabular}{lcc}
\toprule
Pruning order & Post-pruning & After recovery \\
\midrule
Deepest to shallowest (main) & 88.21\% & 88.84\% \\
Shallowest to deepest & 88.04\% & 89.13\% \\
\bottomrule
\end{tabular}
\end{table}
\begin{figure}[t]
    \centering
    \begin{minipage}[t]{0.50\linewidth}
        \centering
        \includegraphics[
            width=\linewidth,
            keepaspectratio
        ]{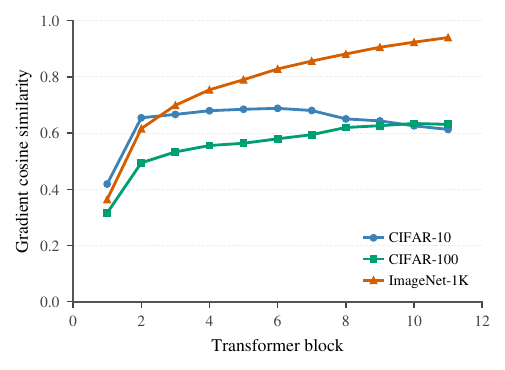}
        \par\smallskip
        {\small (a)}
    \end{minipage}\hfill
    \begin{minipage}[t]{0.46\linewidth}
        \centering
        \includegraphics[
            width=\linewidth,
            keepaspectratio
        ]{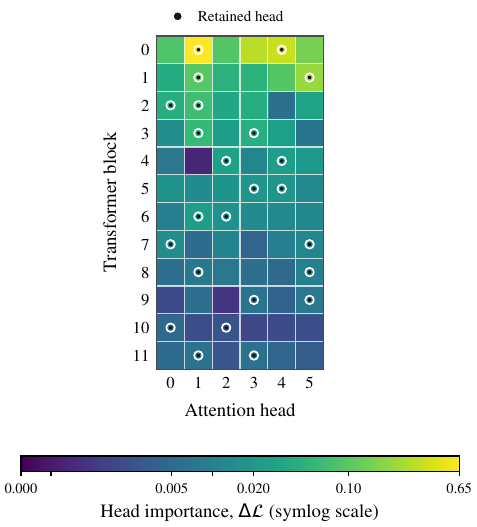}
        \par\smallskip
        {\small (b)}
    \end{minipage}
    \caption{
        \textbf{(a)} Cosine similarity between gradients
        induced by the block-local objective and those
        received by the same block under full backpropagation.
        Local gradients across depth have a mean norm ratio
        of $1.04 \pm 0.02$.
        \textbf{(b)} Each cell shows the increase in final
        classification loss after temporarily ablating one
        attention head. White markers indicate heads retained
        under an example partial-retention budget.
        Best viewed in color.
    }
    \label{fig:gradient-alignment-and-head-importance}
\end{figure}

\subsection{Representation Analysis}
\noindent \textbf{Gradient analysis.} In Figure \ref{fig:gradient-alignment-and-head-importance}(a), we compare the gradients induced by each block-local objective with those induced by the final classification objective, computed with respect to the same block parameters at identical model weights and on identical mini-batches. The cosine similarity remains positive across all blocks and datasets, showing that our local losses favor directions compatible with the end-to-end variant. The alignment is imperfect, particularly on CIFAR-100, indicating that \method models do not receive the complete coordination signal available through full-backpropagation. At the same time, the mean gradient-norm ratio is $1.04 \pm 0.02$, showing that the difference primarily
concerns direction rather than gradient scale.

\noindent \textbf{Pruning score heatmap.} In Figure \ref{fig:gradient-alignment-and-head-importance}(b), we showcase the importance assigned to each attention head.
Head importance is measured as
$s_u^{(i)}
            =
            \max\left(
            0,
            \mathcal{L}^{(i)}_{\mathrm{final},-u}
            -
            \mathcal{L}_{\mathrm{final}}
            \right)$, where
$\mathcal{L}^{(i)}_{\mathrm{final},-u}$ is the final classifier
loss after temporarily suppressing head $u$ in block $i$. The symmetric-log color scale is
linear near zero and logarithmic above $\Delta\mathcal{L}=0.005$,
revealing variation among low-scoring heads, while spanning the complete
observed range from zero to the maximum score.


\noindent \textbf{Pruning-order ablation.} We examine whether sequential ResNet-50 pruning depends on the order in which the layers are processed. Both variants start from the same CIFAR-10-adapted \method checkpoint and use identical target widths, scoring budgets, structured masks, and local recovery process. The deepest-to-shallowest variant processes layers from the output side toward earlier stages, while the reverse variant follows the opposite traversal. The differences are small and indicate that the final recovered performance is largely insensitive to traversal order. 

\begin{figure}[!t]
    \centering

    \includegraphics[
        width=0.60\columnwidth,
        keepaspectratio
    ]{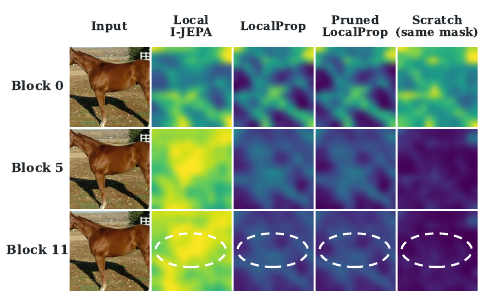}

    \medskip

    \includegraphics[
        width=0.60\columnwidth,
        keepaspectratio
    ]{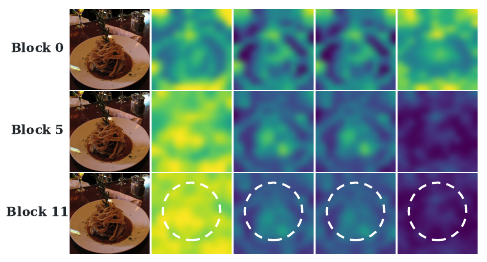}

    \medskip

    \includegraphics[
        width=0.60\columnwidth,
        keepaspectratio
    ]{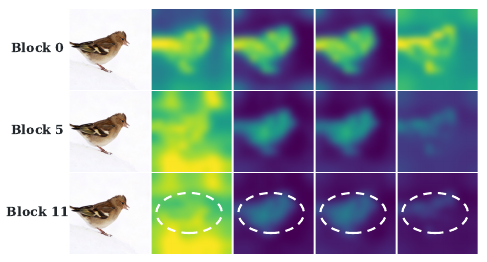}
\vspace{-0.3cm}
    \caption{Additional layerwise feature-magnitude
    visualizations.
    Columns show the input, local I-JEPA, \method, pruned \method and the same-mask architecture trained from scratch. Rows show transformer blocks $0$, $5$, and $11$
    for three ImageNet-1K validation examples. Feature magnitudes are jointly normalized across methods, and dashed contours indicate the same object region. Best viewed in color.}
    \label{fig:images}
\end{figure}

\noindent \textbf{Additional feature visualizations.} In Figure \ref{fig:images}, we extend the analysis from the main paper to three additional ImageNet examples. We display feature-magnitude at different blocks using the same protocol and observe that spatial structure is preserved across different object categories.

\begin{table*}[!t]
\caption{Key hyperparameters for local supervised fine-tuning.
All runs use AdamW and cross-entropy with label smoothing $0.1$.}
\label{tab:finetune_hparams}
\vspace{0.1cm}
\centering
\scriptsize
\setlength{\tabcolsep}{3pt}
\renewcommand{\arraystretch}{0.98}
\begin{tabular}{@{}llcccccc@{}}
\toprule
Model & Dataset(s) & Epochs & Batch size & Learning rate & Weight decay
& LR schedule & Local Updates \\
\midrule

ViT-S/16
& CIFAR-10, CIFAR-100, ImageNet-1K
& 120
& 64
& $5\times10^{-5}$
& $0.05$
& Cosine
& 8 updates / block \\

ResNet-50
& CIFAR-10, CIFAR-100
& 500
& 64
& $3\times10^{-4}$
& $0.01$
& Constant
& all 14 groups \\

ResNet-50
& ImageNet-1K
& 500
& 224
& $3\times10^{-4}$
& $0.01$
& Constant
& all 14 groups \\

\bottomrule
\end{tabular}
\end{table*}

\subsection{Reproducibility Details}

Tables~\ref{tab:ijepa_hparams}, \ref{tab:finetune_hparams}, and
\ref{tab:pruning_hparams} summarize the final configurations used to obtain the results reported in the main paper. 

\begin{table*}[!t]
\caption{Key pruning and recovery hyperparameters.
Structured masks are enforced throughout local recovery by zeroing the corresponding attention projections for ViT-S and convolutional filters
and normalization parameters for ResNet-50.}
\label{tab:pruning_hparams}
\vspace{0.1cm}
\centering
\small
\setlength{\tabcolsep}{1.2pt}
\renewcommand{\arraystretch}{0.96}
\begin{tabular}{@{}lccc@{}}
\toprule
\multirow{2}{*}{Hyperparameter}
& \multirow{2}{*}{ViT-S/16}
& \multicolumn{2}{c}{ResNet-50} \\
\cmidrule(lr){3-4}
& & CIFAR & ImageNet \\
\midrule

Pruning unit
& Attention heads
& Output channels
& Output channels \\

Pruning target
& \shortstack{48/72 heads removed}
& \multicolumn{2}{c}{$1/3$ of each width removed} \\

Scoring budget (batches)
& 8
& 8
& 8\\

Recovery budget (epochs)
& \shortstack{10}
& \shortstack{10}
& \shortstack{10} \\

Recovery batch size
& 64
& 64
& 224 \\

Recovery LR
& $10^{-5}$ / $10^{-4}$
& \shortstack{$10^{-5}$ / $10^{-4}$}
& $10^{-5}$ / $10^{-3}$ \\

Weight decay
& $0.05$
& $10^{-4}$
& $0.05$ \\

Label smoothing
& $0.1$
& $0.1$
& $0.1$ \\

\bottomrule
\end{tabular}
\end{table*}

\begin{table}[!t]
\caption{Top-1 accuracy across three training seeds. Variability is reported using the sample standard deviation.}
\label{tab:cifar_10_seed_sensitivity}
\centering
\small 
\setlength{\tabcolsep}{3pt}
\begin{tabular}{p{2.65cm}cccc}
\toprule
Setting
& Seed 0
& Seed 1
& Seed 42
& $\mu \pm \sigma$ \\
\midrule
\method 
& $76.69$
& $77.17$
& $76.76$
& $76.87 \pm 0.26$ \\

Pruned
& $52.21$
& $53.86$
& $52.36$
& $52.81 \pm 0.91$ \\

Recovered
& $75.40$
& $75.93$
& $75.40$
& $75.58 \pm 0.31$ \\

Same-mask Scratch
& $66.85$
& $66.46$
& $66.68$
& $66.66 \pm 0.20$ \\
\bottomrule
\end{tabular}
\end{table}

\noindent \textbf{Seed sensitivity.}
Table \ref{tab:cifar_10_seed_sensitivity} reports CIFAR-10 results across three seeds. The relative ordering of the configurations remains stable, with recovery restoring most of the pruning loss and outperforming the same-mask model trained from scratch.

\noindent \textbf{Hardware, software, and randomness.}
Experiments ran on NVIDIA RTX 3090 and RTX 5090 GPUs, together with an
Intel Core i9 CPU running Ubuntu 20.04.6. The software environment used
Python 3.8, PyTorch 2.4.1. We seed Python, NumPy, PyTorch, and all CUDA devices before model initialization and data-loader construction. Each main-table entry corresponds to one complete training run. Table \ref{tab:cifar_10_seed_sensitivity} separately reports three-seed results. Full version list is included with the submitted code.  

\noindent \textbf{Evaluation metrics.} We report top-1 classification accuracy as the primary downstream performance metric. Efficiency is measured in the main paper using active parameter count, peak allocated GPU memory, and GPU-hours.

\noindent \textbf{Code and model availability.}
The supplementary code archive contains implementations and configuration files required to reproduce local I-JEPA pre-training, local supervised fine-tuning, pruning and recovery for both architectures. Upon publication, the source code and trained
models will be released under the CC BY-NC 4.0 license.  

\end{document}

%% file: math_commands.tex
\usepackage{amsmath,amsfonts,bm}

\def\eqref#1{equation~\ref{#1}}

\def\1{\bm{1}}

\DeclareMathAlphabet{\mathsfit}{\encodingdefault}{\sfdefault}{m}{sl}
\SetMathAlphabet{\mathsfit}{bold}{\encodingdefault}{\sfdefault}{bx}{n}



%% file: iclr2027_conference.bib
@inproceedings{madan2024cl,
  title="{CL-MAE: Curriculum-learned masked autoencoders}",
  author={Madan, Neelu and Ristea, Nicolae-C{\u{a}}t{\u{a}}lin and Nasrollahi, Kamal and Moeslund, Thomas B and Ionescu, Radu Tudor},
  booktitle={Proceedings of WACV},
  pages={2480--2490},
  year={2024},
}

@inproceedings{grigore2025weight,
  title={Weight copy and low-rank adaptation for few-shot distillation of vision transformers},
  author={Grigore, Diana-Nicoleta and Georgescu, Mariana-Iuliana and Justo, Jon Alvarez and Johansen, Tor and Ionescu, Andreea Iuliana and Ionescu, Radu Tudor},
  booktitle={Proceedings of WACV},
  pages={7368--7378},
  year={2025},
}

@article{hondru2025masked,
  title={Masked image modeling: A survey},
  author={Hondru, Vlad and Croitoru, Florinel Alin and Minaee, Shervin and Ionescu, Radu Tudor and Sebe, Nicu},
  journal={International Journal of Computer Vision},
  volume={133},
  number={10},
  pages={7154--7200},
  year={2025},
}

@article{hinton2022forward,
  title         = "{The Forward-Forward Algorithm: Some Preliminary Investigations}",
  author        = {Hinton, Geoffrey},
  year          = {2022},
  journal={arXiv preprint arXiv:2212.13345},
}

@inproceedings{li2026noprop,
  title={Noprop: Training neural networks without back-propagation or forward-propagation},
  author={Li, Qinyu and Teh, Yee Whye and Pascanu, Razvan},
  booktitle={Proceedings of CoLLAs},
  pages={525--544},
  year={2026},
}

@article{rumelhart1986learning,
  title   = {Learning representations by back-propagating errors},
  author  = {Rumelhart, David E. and Hinton, Geoffrey E. and Williams, Ronald J.},
  journal = {Nature},
  volume  = {323},
  pages   = {533--536},
  year    = {1986}
}

@article{lillicrap2016random,
  title   = {Random synaptic feedback weights support error backpropagation for deep learning},
  author  = {Lillicrap, Timothy P. and Cownden, Daniel and Tweed, Douglas B. and Akerman, Colin J.},
  journal = {Nature Communications},
  volume  = {7},
  number  = {13276},
  year    = {2016}
}

@article{siddiqui2024blockwise,
  title={Blockwise Self-Supervised Learning at Scale},
  author={Siddiqui, Shoaib Ahmed and Krueger, David and
          LeCun, Yann and Deny, St{\'e}phane},
  journal={Transactions on Machine Learning Research},
  year={2024},
}

@inproceedings{ma2024scaling,
  title={Scaling Supervised Local Learning with Augmented Auxiliary Networks},
  author={Ma, Chenxiang and Wu, Jibin and Si, Chenyang and
          Tan, Kay Chen},
  booktitle={Proceedings of ICLR},
  year={2024},
}

@article{lillicrap2020backprop,
  title   = {Backpropagation and the brain},
  author  = {Lillicrap, Timothy P. and Santoro, Adam and Marris, Luke and Akerman, Colin J. and Hinton, Geoffrey},
  journal = {Nature Reviews Neuroscience},
  volume  = {21},
  pages   = {335--346},
  year    = {2020}
}

@inproceedings{lee2015difference,
  title     = {Difference Target Propagation},
  author    = {Lee, Dong-Hyun and Zhang, Saizheng and Fischer, Asja and Bengio, Yoshua},
  booktitle = {Proceedings of ECML-PKDD},
  pages     = {498--515},
  year      = {2015}
}

@inproceedings{nokland2019training,
  title     = {Training Neural Networks with Local Error Signals},
  author    = {N{\o}kland, Arild and Eidnes, Lars Hiller},
  booktitle = {Proceedings of ICML},
  pages     = {4839--4850},
  year      = {2019}
}

@inproceedings{belilovsky2020decoupled,
  title     = {Decoupled Greedy Learning of {CNN}s},
  author    = {Belilovsky, Eugene and Eickenberg, Michael and Oyallon, Edouard},
  booktitle = {Proceedings of ICML},
  volume    = {119},
  pages     = {736--745},
  year      = {2020}
}

@inproceedings{pyeon2021sedona,
  title     = {{SEDONA}: Search for Decoupled Neural Networks toward Greedy Block-wise Learning},
  author    = {Pyeon, Myeongjang and Moon, Jihwan and Hahn, Taeyoung and Kim, Gunhee},
  booktitle = {Proceedings of ICLR},
  year      = {2021}
}

@article{hinton2015distilling,
  title   = {Distilling the Knowledge in a Neural Network},
  author  = {Hinton, Geoffrey and Vinyals, Oriol and Dean, Jeff},
  journal = {arXiv preprint arXiv:1503.02531},
  year    = {2015}
}

@inproceedings{zhao2023cumulative,
  title={Cumulative spatial knowledge distillation for vision transformers},
  author={Zhao, Borui and Song, Renjie and Liang, Jiajun},
  booktitle={Proceedings of ICCV},
  pages={6146--6155},
  year={2023}
}

@inproceedings{romero2015fitnets,
  title     = "{FitNets: Hints for Thin Deep Nets}",
  author    = {Romero, Adriana and Ballas, Nicolas and Kahou, Samira Ebrahimi
               and Chassang, Antoine and Gatta, Carlo and Bengio, Yoshua},
  booktitle = {Proceedings of ICLR},
  year      = {2015}
}

@inproceedings{zhang2019byot,
  title     = {Be Your Own Teacher: Improve the Performance of Convolutional
               Neural Networks via Self Distillation},
  author    = {Zhang, Linfeng and Song, Jiebo and Gao, Anni and Chen, Jingwei
               and Bao, Chenglong and Ma, Kaisheng},
  booktitle = {Proceedings of ICCV},
  pages     = {3713--3722},
  year      = {2019}
}

@inproceedings{Dosovitskiy2021ViT,
  title     = {An Image is Worth 16x16 Words: Transformers for Image Recognition at Scale},
  author    = {Dosovitskiy, Alexey and Beyer, Lucas and Kolesnikov, Alexander and Weissenborn, Dirk and Zhai, Xiaohua and Unterthiner, Thomas and Dehghani, Mostafa and Minderer, Matthias and Heigold, Georg and Gelly, Sylvain and Uszkoreit, Jakob and Houlsby, Neil},
  booktitle = {Proceedings of ICLR},
  year      = {2021}
}

@InProceedings{Liu_2021_ICCV,
  title     = {Image Retrieval on Real-Life Images With Pre-Trained Vision-and-Language Models},
  author    = {Liu, Zheyuan and Rodriguez-Opazo, Cristian and Teney, Damien and Gould, Stephen},
  booktitle = {Proceedings of ICCV},
  month     = {October},
  year      = {2021},
  pages     = {2125-2134}
  }

@inproceedings{carion2020end,
  title={End-to-end object detection with transformers},
  author={Carion, Nicolas and Massa, Francisco and Synnaeve, Gabriel and Usunier, Nicolas and Kirillov, Alexander and Zagoruyko, Sergey},
  booktitle={Proceedings of ECCV},
  pages={213--229},
  year={2020},
}

@inproceedings{kirillov2023segment,
  title={Segment anything},
  author={Kirillov, Alexander and Mintun, Eric and Ravi, Nikhila and Mao, Hanzi and Rolland, Chloe and Gustafson, Laura and Xiao, Tete and Whitehead, Spencer and Berg, Alexander C and Lo, Wan-Yen and others},
  booktitle={Proceedings of ICCV},
  pages={4015--4026},
  year={2023}
}

@inproceedings{mildenhall2020nerf,
  title="{NeRF: Representing scenes as neural radiance fields for view synthesis}",
  author={Mildenhall, Ben and Srinivasan, Pratul P. and Tancik, Matthew and Barron, Jonathan T. and Ramamoorthi, Ravi and Ng, Ren},
  booktitle={Proceedings of ECCV},
  pages={405--421},
  year={2020},
}

@inproceedings{he2022masked,
  title={Masked autoencoders are scalable vision learners},
  author={He, Kaiming and Chen, Xinlei and Xie, Saining and Li, Yanghao and Doll{\'a}r, Piotr and Girshick, Ross},
  booktitle={Proceedings of CVPR},
  pages={16000--16009},
  year={2022}
}

@inproceedings{chen2020simple,
  title={A simple framework for contrastive learning of visual representations},
  author={Chen, Ting and Kornblith, Simon and Norouzi, Mohammad and Hinton, Geoffrey},
  booktitle={Proceedings of ICML},
  pages={1597--1607},
  year={2020},
}

@inproceedings{han2015learning,
  title={Learning both weights and connections for efficient neural network},
  author={Han, Song and Pool, Jeff and Tran, John and Dally, William},
  booktitle={Proceedings of NeurIPS},
  volume={28},
  year={2015}
}

@inproceedings{li2016pruning,
  title={Pruning filters for efficient convnets},
  author={Li, Hao and Kadav, Asim and Durdanovic, Igor and Samet, Hanan and Graf, Hans Peter},
  booktitle={Proceedings of ICLR},
  year={2017}
}

@article{chen2016training,
  title={Training deep nets with sublinear memory cost},
  author={Chen, Tianqi and Xu, Bing and Zhang, Chiyuan and Guestrin, Carlos},
  journal={arXiv preprint arXiv:1604.06174},
  year={2016}
}

@article{wright2025distinct,
  title={Distinct synaptic plasticity rules operate across dendritic compartments in vivo during learning},
  author={Wright, William J. and Hedrick, Nathan G. and Komiyama, Takaki},
  journal={Science},
  volume={388},
  year={2025},
  publisher={American Association for the Advancement of Science}
}

@article{hedrick2024local,
  title={Local and global predictors of synapse elimination during motor learning},
  author={Hedrick, Nathan G. and Wright, William J. and Komiyama, Takaki},
  journal={Science Advances},
  volume={10},
  number={11},
  year={2024},
  publisher={American Association for the Advancement of Science}
}

@inproceedings{he2020moco,
  title     = {Momentum Contrast for Unsupervised Visual Representation Learning},
  author    = {He, Kaiming and Fan, Haoqi and Wu, Yuxin and Xie, Saining
               and Girshick, Ross},
  booktitle = {Proceedings of CVPR},
  pages     = {9729--9738},
  year      = {2020}
}

@inproceedings{grill2020byol,
  title     = {Bootstrap Your Own Latent: A New Approach to Self-Supervised Learning},
  author    = {Grill, Jean-Bastien and Strub, Florian and Altch{\'e}, Florent
               and Tallec, Corentin and Richemond, Pierre H. and Buchatskaya, Elena
               and Doersch, Carl and Pires, Bernardo Avila and Guo, Zhaohan Daniel
               and Azar, Mohammad Gheshlaghi and Piot, Bilal
               and Kavukcuoglu, Koray and Munos, R{\'e}mi and Valko, Michal},
  booktitle = {Proceedings of NeurIPS},
  volume    = {33},
  year      = {2020}
}

@inproceedings{caron2021dino,
  title     = {Emerging Properties in Self-Supervised Vision Transformers},
  author    = {Caron, Mathilde and Touvron, Hugo and Misra, Ishan
               and J{\'e}gou, Herv{\'e} and Mairal, Julien
               and Bojanowski, Piotr and Joulin, Armand},
  booktitle = {Proceedings of ICCV},
  pages     = {9650--9660},
  year      = {2021}
}

@article{lecun2015deep,
  title={Deep learning},
  author={LeCun, Yann and Bengio, Yoshua and Hinton, Geoffrey},
  journal={Nature},
  volume={521},
  number={7553},
  pages={436--444},
  year={2015}
}

@article{schmidhuber2015deep,
  title={Deep learning in neural networks: An overview},
  author={Schmidhuber, J{\"u}rgen},
  journal={Neural Networks},
  volume={61},
  pages={85--117},
  year={2015}
}

@inproceedings{assran2023self,
  title={Self-supervised learning from images with a joint-embedding predictive architecture},
  author={Assran, Mahmoud and Duval, Quentin and Misra, Ishan and Bojanowski, Piotr and Vincent, Pascal and Rabbat, Michael and LeCun, Yann and Ballas, Nicolas},
  booktitle={Proceedings of CVPR},
  pages={15619--15629},
  year={2023}
}

@inproceedings{molchanov2019importance,
  title={Importance estimation for neural network pruning},
  author={Molchanov, Pavlo and Mallya, Arun and Tyree, Stephen and Frosio, Iuri and Kautz, Jan},
  booktitle={Proceedings of CVPR},
  pages={11264--11272},
  year={2019}
}

@inproceedings{fang2023depgraph,
  title={Depgraph: Towards any structural pruning},
  author={Fang, Gongfan and Ma, Xinyin and Song, Mingli and Mi, Michael Bi and Wang, Xinchao},
  booktitle={Proceedings of CVPR},
  pages={16091--16101},
  year={2023}
}

@inproceedings{yu2022unified,
  title={A unified framework for compression and acceleration of vision transformers},
  author={Yu, Shixing and Zhao, Tianlong and Yin, Jiwen and Liu, Hui and Yang, Jian},
  booktitle={Proceedings of CVPR},
  pages={12124--12133},
  year={2022}
}

@inproceedings{papac2024convolutional,
  title={Convolutional channel-wise competitive learning for the forward-forward algorithm},
  author={Papachristodoulou, Andreas and Kyrkou, Christos and Timotheou, Stelios and Theocharides, Theocharis},
  booktitle={Proceedings of AAAI},
  volume={38},
  pages={14536--14544},
  year={2024}
}

@article{sarode2026hyperspherical,
  title={Hyperspherical Forward-Forward with Prototypical Representations},
  author={Sarode, Shalini and Moser, Brian and Folz, Joachim and Raue, Federico and Nauen, Tobias and Frolov, Stanislav and Dengel, Andreas},
  journal={arXiv preprint arXiv:2605.00082},
  year={2026}
}

@techreport{krizhevsky2009learning,
  title={Learning multiple layers of features from tiny images},
  author={Krizhevsky, Alex and Hinton, Geoffrey},
  year={2009},
  institution={Department of Computer Science, University of Toronto}
}

@inproceedings{deng2009imagenet,
  title="{ImageNet: A large-scale hierarchical image database}",
  author={Deng, Jia and Dong, Wei and Socher, Richard and Li, Li-Jia and Li, Kai and Fei-Fei, Li},
  booktitle={Proceedings of CVPR},
  pages={248--255},
  year={2009}
}

@misc{lecun2010mnist,
  title={MNIST handwritten digit database},
  author={LeCun, Yann and Cortes, Corinna and Burges, Christopher JC},
  year={2010},
  howpublished={Available: \url{http://yann.lecun.com/exdb/mnist/}}
}

@inproceedings{kalapos2024cnnjepa,
  title     = "{CNN-JEPA: Self-Supervised Pretraining Convolutional Neural Networks Using Joint Embedding Predictive Architecture}",
  author    = {Kalapos, Andr{\'a}s and Gyires-T{\'o}th, B{\'a}lint},
  booktitle = {Proceedings of ICMLA},
  pages     = {1111--1114},
  year      = {2024},
}

@inproceedings{lowe2019end,
  title={Putting An End to End-to-End: Gradient-Isolated Learning of Representations},
  author={L{\"o}we, Sindy and O'Connor, Peter and Veeling, Bastiaan S.},
  booktitle={Proceedings of NeurIPS},
  volume={32},
  year={2019}
}

@inproceedings{yosinski2014transferable,
  title={How transferable are features in deep neural networks?},
  author={Yosinski, Jason and Clune, Jeff and Bengio, Yoshua and Lipson, Hod},
  booktitle={Proceedings of NeurIPS},
  volume={27},
  year={2014}
  }

@inproceedings{chen2022when,
  title="{When Vision Transformers Outperform ResNets without Pre-Training or Strong Data Augmentations}",
  author={Chen, Xiangning and Hsieh, Cho-Jui and Gong, Boqing},
  booktitle={Proceedings of ICLR},
  year={2022}
}

@inproceedings{mo2024connecting,
  title={Connecting Joint-Embedding Predictive Architecture with Contrastive Self-Supervised Learning},
  author={Mo, Shentong and Tong, Shengbang},
  booktitle={Proceedings of NeurIPS},
  volume={37},
  year={2024}
}

@misc{torchvision_resnet50,
  author={{TorchVision Contributors}},
  title={{ResNet-50} Pretrained Weights},
  year={2022},
  note={{IMAGENET1K\_V1}: 76.13\% ImageNet-1K top-1 accuracy}
}
